\documentclass[journal]{IEEEtran}

\ifCLASSINFOpdf
\else
\fi

\usepackage{algorithm}
\usepackage{algpseudocode}
\usepackage{graphicx}
\usepackage{amssymb}
\usepackage{amssymb,mathtools}
\usepackage{algpseudocode}
\usepackage{amsmath}
\usepackage{hyperref}
\usepackage{pslatex}
\usepackage{float}
\usepackage{xcolor}
\usepackage[normalem]{ulem}

\newcommand{\GV}[1]{\textcolor{black}{#1}}
\newcommand{\GVnew}[1]{\textcolor{black}{#1}}

\begin{document}

\title{\GV{A Deep Learning Technique to Control the Non-linear Dynamics of a Gravitational-wave Interferometer.}}

\author{Peter Xiangyuan Ma, Department of Mathematics, University of Toronto
        
Gabriele Vajente, LIGO Laboratory, California Institute of Technology

\thanks{P. Ma is with the Department of Mathematics, University of Toronto, 40 St. George Street, Toronto, ON M5S 2E4, Canada e-mail: (peterxy.ma@mail.utoronto.ca)

G. Vajente is with the LIGO Laboratory at the California Institute of Technology, 1200 E. California Blvd, Pasadena (CA) 91101 email: (vajente@caltech.edu)}% <-this % stops a space
}

\markboth{PREPRINT, Febuary~2023}%
{Ma \MakeLowercase{\textit{et al.}}: Bare Demo of IEEEtran.cls for IEEE Journals}

\maketitle

\begin{abstract}
\GV{In this work we developed a deep learning technique that successfully solves a non-linear dynamic control problem. Instead of directly tackling the control problem, we combined methods in probabilistic neural networks and a Kalman-Filter-inspired model to build a non-linear state estimator for the system. We then used the estimated states to implement a trivial controller for the now fully observable system. We applied this technique to a crucial non-linear control problem that arises in the operation of the LIGO system, an interferometric gravitational-wave observatory. We demonstrated in simulation that our approach can learn from data to estimate the state of the system, allowing a successful control of the interferometer's mirror . We also developed a computationally efficient model that can run in real time at high sampling rate on a single modern CPU core, one of the key requirements for the implementation of our solution in the LIGO digital control system. We believe these techniques could be used to help tackle similar non-linear control problems in other applications. }
\end{abstract}


\IEEEpeerreviewmaketitle

\section{Introduction}
\subsection{Non-linear Dynamic Controls}

\GV{Some of the hardest problems faced in the area of control theory deal with non-linear control problems. While a well established theoretical and practical formalism exists to design feedback controllers for linear systems, there is no general approach that work in the non-linear case \cite{nonlinear_dynamics_intro}. Ad-hoc approaches are often developed on a case-by-case basis. Non-linear problems like the classic inverted pendulum have well known and studied solutions \cite{inverted_pendulum}. Other commonly used approaches involve developing approximations to recast the non-linear problem into a linear one by local mapping of the observables or states \cite{control_survey_nonlinear}. Often the solutions involve cleverly addressing specific problems by making certain simplifying assumptions about the system. }

\GV{Recently, due to the advent of machine learning and deep learning we are seeing new, more generally applicable approaches. Deep Neural Networks offer the promise of efficiently fitting complex non-linear relationships and are therefore considered a potential basis for novel non-linear control strategies. Research in areas like deep reinforcement learning are developing algorithms that learn control policies directly from interacting with these complex systems from scratch \cite{rl_nonlinear}.  }

Overall, one of the goals in the field of control theory is to explore how these novel deep learning based techniques can address specific non-linear control problems and to gauge their performance against classical techniques. With this motivation, we explore such a case in this paper. We study how deep learning can solve \GV{the problem of controlling the longitudinal translational degrees of freedom of the LIGO (Laser Interferometer Gravitational-Wave Observatory) detector, bringing the system from a condition where all degrees of freedom are varying over a large phase space, to a situation when the system is tightly controlled around the operating point, where the system is highly linear. The only observables available to the control system are highly non-linear function of the state. This \emph{lock acquisition} problem \cite{pdh}\cite{trad_lock_2} has been historically solved with solutions developed case by case that are difficult to scale to more complex systems. Deep Learning techniques have also been recently applied to the steady state linear control of angular motions in gravitational-wave interferometers \cite{rl_ligo_arxiv}.}

\subsection{Introduction to LIGO}
\label{sec:what_is_ligo}
\GV{The LIGO (Laser Interferometer Gravitational-wave Observatory) is a 4-km-long Michelson laser interferometer \cite{LIGO_textbook}, with the goal of detecting the very small differential distance variation (of the order of $10^{-21}$ m and smaller) caused by the space-time metric fluctuation generated by the passage of gravitational waves \cite{gw-einstein} produced by astrophysical events \cite{original_gw_detect}. To reach such extreme sensitivity the LIGO detectors employ multiple resonant optical cavities \cite{Vajente_textbook}, that need to be maintained at resonance using feedback control systems: high sensitivity and a linear response of the optical system is obtained only when the distances between the mirrors composing the cavities are kept close to the operating point.  Additionally, to isolate the system from ground vibration, all mirrors are suspended to sophisticated seismic isolation systems \cite{sus, sei}: the mirrors motion at the frequency of interest for the detection of gravitational waves (above 10 Hz) is largely reduced, at the price of a large residual motion at low frequency that spans multiple resonances of the optical cavities. LIGO uses a frontal modulation technique \cite{ LIGO_textbook, pdh} to extract optical signals that can be used to precisely measure the fluctuation of all length degrees of freedom from the operating point. Those signals, readout by photo-detectors probing pick-off laser beams at different places in the instrument,  provide a highly sensitive observation of the distance between mirrors (the state of the optical system we are interested in) only for a small region of the phase space, near the operating point. For most of the state space spanned by the seismically-induced random motion of the mirrors the signal responses are highly non linear.}

\begin{figure}[H]
\centering
\includegraphics[width=\linewidth]{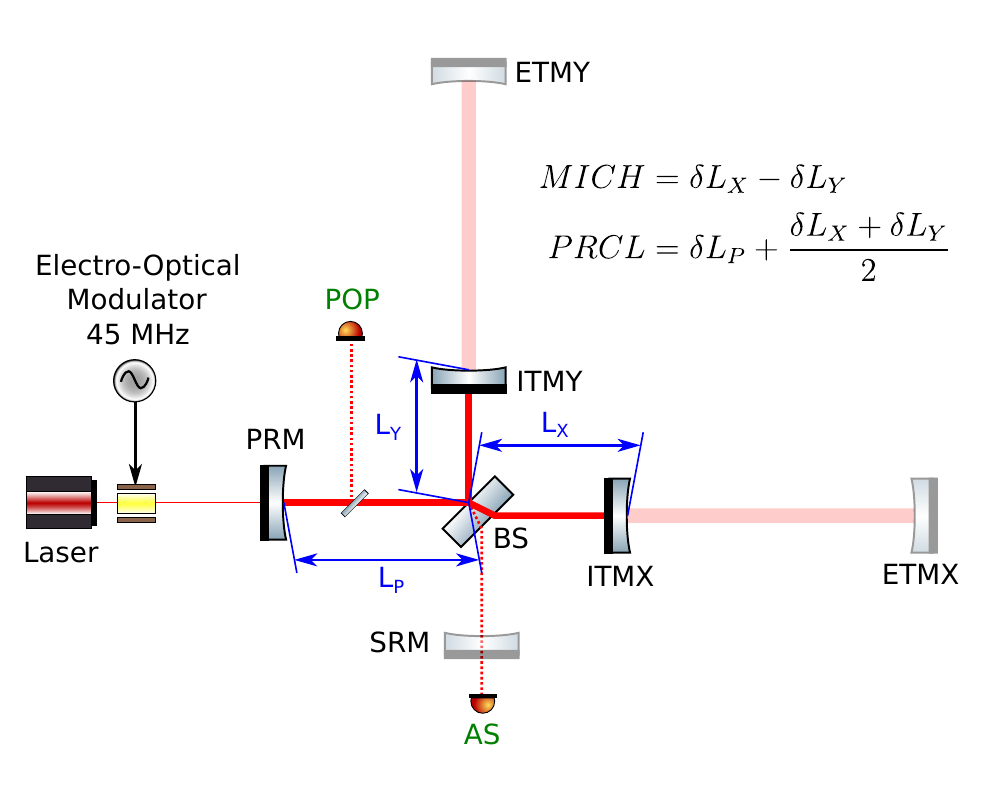}
\caption{\label{fig:ligo_mirror} \GV{A simplified optical layout of the full Advanced LIGO interferometer, not to scale. The input laser is phase modulated at several RF frequencies, including 45 MHz, to allow implementing a Pound-Drever-Hall sensing scheme \cite{pdh}. The two 4-km-long Fabry-Perot resonant cavities in the arms are comprised of Input Test Masses (ITMX and ITMY) ans End Test Masses (ETMX and ETMY). The input laser beam is separated in two equal parts by the Beam Splitter (BS) mirror. The additional Power Recycling Mirror (PRM) in the input path amplifies the laser power circulating inside the interferometer, and the output Signal Recycling Mirror (SRM) is used to increase the detector response to gravitational waves. For the study considered here, only a reduced optical configuration is considered, composed of the PRM, BS, ITMX and ITMY. The other mirrors, greyed out in the layout, are misaligned so that they do not contribute to the optical system behavior. Two longitudinal degrees of freedom, MICH and PRCL, are relevant for this configuration, and are defined with the equation in the figure, from the microscopic changes of the mirror relative distances, measured from the operating point. Two photo detectors are probing laser beams (POP, a pick-off of the Power Recycling Cavity and AS, the anti-symmetric transmission into the Signal Recycling Cavity)  and providing the observable optical signals.} }
\end{figure}

In order to \GV{further enhance the detector sensitivity to gravitational waves}, LIGO uses \GV{additional resonant cavities to increase the laser power circulating inside the interferometer and to shape the systems response to gravitational waves} \cite{intro_ligo}. \GV{The additional Power Recycling and Signal Recycling mirrors (see figure \ref{fig:ligo_mirror}) add complexity to the control problem.} %One of these critical systems is the PRMI (Power Recycling Michelson Interferometer) \cite{Vajente_textbook}. This set of mirrors includes the power recycling mirrors along with the input test mass. The purpose of the Power Recycling Mirror is that once we use the mirrors to tune the laser into resonance it puts more power through the entire interferometer \cite{LIGO_textbook}. A more powerful laser helps improve the sensitivity of the detector. The input test masses are the mirrors right before entering the arms of the detector where Fabry-Perot cavities. Please see the figure \ref{fig:ligo_mirror} for the graphical depiction of the setup.

\subsection{Challenges}
\label{sec:general_challenges}
The task \GV{of bringing the longitudinal degrees of freedom of a gravitational wave interferometric detector from the initial state, dominated by large random motion, to the final high sensitivity configuration, where all degrees of freedom are kept very close to the working point, is called \emph{lock acquisition} \cite{trad_lock_2, trad_lock_1}. The controllers use the optical signals as inputs, and are able to apply forces to the mirrors through suitable actuators in the seismic isolation and suspension chain \cite{ sus, sei}.}

Traditionally, controlling the motion of a pendulum is a well known and solved problem \cite{pendulum}. However, these methods require information about the \textit{state} of the pendulum, \GV{that in our case corresponds to the distances between mirrors}. In our problem \GV{we do not have the capability of observing continuously the states with the accuracy needed to control the system}. The only data we can retrieve from the system are the optical signals, which are \textit{non-linear} and \textit{non-unique} \GV{functions} of the mirror relative positions \cite{Vajente_textbook}. \GV{In other words, the optical signals are linearly related to the systems state only for very small fluctuations around the working point: once the system has been driven in this linear regime, classical linear feedback controllers can be used to maintain the lock. However, this linear region is a small fraction, of the order of $10^{-6}$ or smaller of the entire explorable state space: in most of the state space the optical signals are non-linear and non-invertible functions of the state variables.} \GV{The non-uniqueness rise from the periodic nature of the laser wave: any change in the state that increases or decreases the distance travelled by laser beams by multiple of the wavelength has no effect on the optical signals. This also implies that there is no unique working point that is acceptable: any shift that meets the criterion specified above provides us with an equally suitable working point.}

Hence this is a difficult control problem, where the system dynamics is linear, but the observables are non-linear functions of the states. Despite the difficulty,  as long as we can \GV{estimate} the  state of the system, that is the distances between mirrors, the control problem is solvable by classical \GV{linear} techniques. Thus the central problem is to construct a non-linear state estimator \GV{based on the measurable optical signals} which we can then \GV{be the input} to a simple control technique to acquire the lock.

\subsection{Current Lock Acquisition Solutions}\label{sec:status_quo}

\GV{The entire Advanced LIGO interferometric detector has five main longitudinal degrees of freedom (see figure \ref{fig:ligo_mirror}), with seismically-induced fluctuations that without any control can span several laser wavelengths. To simplify the lock acquisition problem, auxiliary lasers with a different wavelength are injected from the end of the arm resonant cavities \cite{Staley2014}, allowing independent control of the two 4-km-long cavities. The remaining interferometer degrees of freedom are three (see figure \ref{fig:ligo_mirror}): the differential distance between the beamsplitter and the two input test masses (MICH); the average distance between the beam splitter and the two test masses plus the distance from the beam splitter to the power recyling mirror (PRCL); the average distance between the beam splitter and the two test masses plus the distance from the beam splitter to the signal recyling mirror (SRCL). The lock acquisition of those three degrees of freedom works} by attempting to switch on classical feedback controls when the mirrors pass through certain relative resonance conditions, \GV{determined by power level crossings in several photodiode signals \cite{drmi-lock}}. Although this technique works in the current LIGO detectors, it is not ideal: \GV{the intrinsic randomness of the mirror motion makes it impossible to predict how long it would take for the lock acquisition, and the process can sometimes last tens of minutes; development of the technique} involves expert knowledge about the system; when building new \GV{and more complex} detectors this technique is not directly transferable.

\GV{In this work we study the possibility of applying a Machine Learning technique to solve the problem of the lock acquisition of the corner degrees of freedom. For simplicity, we study the case of a Power Recycled Michelson (PRMI) interferometer, where the signal recycling mirror is omitted. Although this configuration is simpler than the Dual Recycled Michelson (DRMI) case used in Advanced LIGO, it contains the characteristic non-linearity and non-uniqueness of the full problem, and therefore it serves as a suitable proof of principle of the main ideas.}

\subsection{Non-Linear and Non-Uniqueness of Optical Signals}
\label{sec:non_linear}
Before attempting to solve the control problem we need to understand what makes the relationship between the optical signals and the state of the mirrors non-linear and non-unique. \GV{The optical signals are obtained from the laser fields exiting from various ports of the interferometers. Following an extension of the Pound-Drever-Hall technique called frontal modulation, the input laser beam is phase-modulated at fixed radio-frequencies. The laser fields are probed by fast photodetectors and the output signal demodulated so to obtain the full set of signals used to observe the state \cite{frontal}. The frontal modulation produces additional sidebands fields spaced around the main laser field by the modulation frequency $f_{mod}$ which is typically of the order of several tens of MHz. The carrier and sideband fields propagate inside the interferometer independently, and are mixed only at the photodetector level.}

In \GV{the} PRMI setup \GV{we are considering for this study}, \GV{there are} two important \GV{longitudinal degrees of freedom} at play: PRCL and MICH. \GV{We consider the zero of both degrees of freedom to be the operating point, so that when propagating inside the PRMI the laser fields accumulate a phase given by},

\begin{equation}
\phi_{MICH} = k\delta L_{MICH}\pm \frac{\Omega}{c}L_{MICH} \\ 
\label{eq:phimich}
\end{equation}
\begin{equation}
\phi_{PRC} = k\delta L_{PRC}\pm \frac{\Omega}{c}L_{PRC} \\ 
\label{eq:phiprcl}
\end{equation}
were $k=2\pi/\lambda$ is \GV{related to the laser wavelenght $\lambda$} and \(\Omega = 2\pi\ f_{mod}\) \cite{Vajente_textbook}. Note that \(\delta L_{MICH}\) and \(\delta L_{PRCL}\) are the deviation of MICH and PRCL from the operating point

\GV{An analytical expression for the laser fields in the reflection, power recycling cavity, anti-symmetric port and anti-symmetric ports of the interfrometer, can be derived following methods similar to what is explained in \cite{Vajente_textbook}. Here we report the results, since the derivation is outside the scope of this paper. Note that  \(\Psi_{IN}\) is the input laser field (either carrier or sidebands)}

\begin{equation}
r_{MICH} = r_X t^2_{BS}e^{i\phi_{MICH}} + r_Y r^2_{BS}e^{-i\phi_{MICH}}  
\label{eq:rmich}
\end{equation}
\begin{equation}
t_{MICH} = t_{BS}r_{BS}(r_Xe^{i\phi_{MICH}} - r_Ye^{-i\phi_{MICH}} )
\label{eq:tmich}
\end{equation}
\begin{equation}
\Psi_{PRC} = \frac{t_{PR}}{1 - r_{PR}}r_{MICH}e^{2i\phi_{PRC}} \Psi_{IN}
\label{eq:field1}
\end{equation}
\begin{equation}
\Psi_{REF} = \frac{i r_{PR} - it^2_{PR}r_{MICH}e^{2i\phi_{PRC}}}{1- r_{PR}r_{MICH}e^{2i\phi_{PRC}}} \Psi_{IN}
\label{eq:field2}
\end{equation}
\begin{equation}
\Psi_{AP} = \frac{it_{MICH}t_{PR}e^{i\phi_{PRC}}}{1-r_{PR}r_{MICH}e^{2i\phi_{PRC}}} \Psi_{IN}
\label{eq:field3}
\end{equation}
\GV{Those equations can be used to compute the main carrier laser fields, for which $\Omega=0$ as well as the two sideband fields with $\Omega = \pm 2\pi f_{mod}$. The optical signals are obtained as combination of the product of pair of fields, from photodetectors that can measured the time-varying power impinging on them}:
\begin{equation}
P(t) = \left| \Psi_0(t) + e^{i\Omega t}\Psi_{+\Omega}(t) + e^{-i\Omega t}\Psi_{-\Omega}(t)\right|^2
\end{equation}
\GV{The low frequency component (well below $f_{mod}$) is called the DC component. The power $P(t)$ contains also a component oscillating at the modulation frequency $f_{mod}$: the real and imaginary parts of this component can be extracted by demodulating the photodetector output with a heterodyne scheme and provide the I and Q quadrature signals. All signals are products of pairs of field amplitudes and complex conjugates.}

\GV{Inspection of equations} \ref{eq:field1},\ref{eq:field2}, \ref{eq:field3} \GV{shows} that the fields contain \GV{exponential} terms that are functions of the MICH and PRCL \GV{degrees of freedom}, \GV{and} thus are periodic. %Since these are periodic functions we know they must be unique only up to a integer multiple of some wavelength and hence motivates both the non-linear and non-unique element of the problem. 
\GV{The presence of those exponential terms explains the fact that multiple positions, differing by multiples of $\lambda/2$, result in the same values for the observed signals, and are therefore indistinguishable and equally suitable as operating point. Additionally, the exponential terms, together with the fact that the optical signals are products of pairs of fields, explain the highly non-linear dependency on MICH and PRCL.}

\subsection{Numerical Simulation}
\label{sec:simulate}
\begin{figure}[tb]
\includegraphics[width=0.95\linewidth]{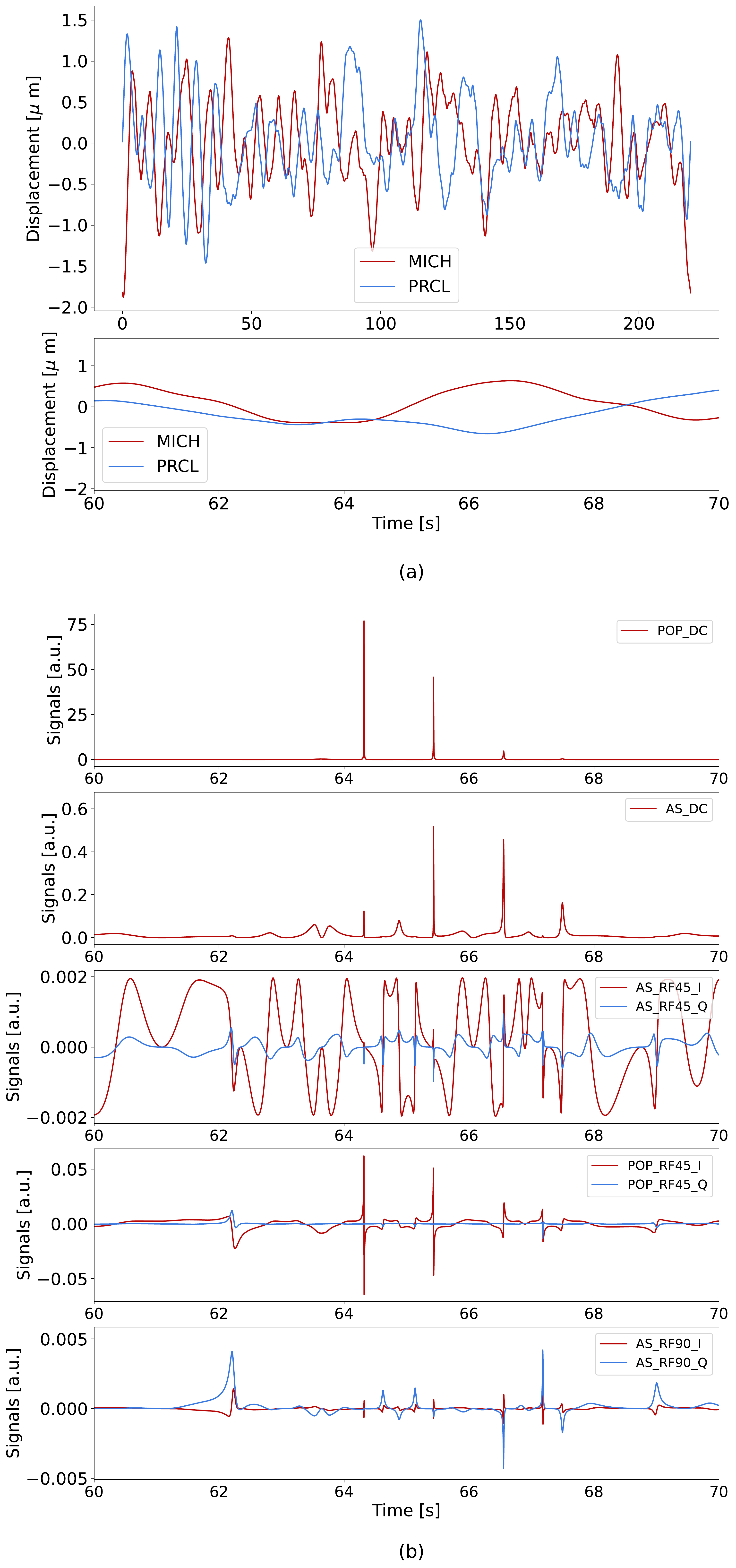}
\label{fig:motion_signals}
\caption{ (a) Shows the simulated 120 seconds worth of trajectories of PRCL/ MICH. (b) is a zoomed-in plot of the optical signals. For naming conventions, figure \ref{fig:ligo_mirror}: AS is the antisymmetric port, POP is the PRC pick-off. For each of those beam, we sense the total power (labelled 'DC'), the Pound-Drever-Hall demodulated signals at 45 MHz (labelled 'RF45 I' and 'RF45 Q') and at 90 MHz (labelled 'RF90 I' and 'RF90 Q'). Each of the demodulated signals has two quadratures: in-phase (I) and quadrature (Q) referring to the fact that we look for a RF signal at the photo-diode that is either in-phase with the laser frontal modulation, or out-of-phase by 90 degrees.} 
\end{figure}

\GV{Our work is based on numerical simulations of the interferometer motion, obtained by creating time series of the MICH and PRCL degrees of freedom that mimic what is observed in the real world.} Then using the field equations \ref{eq:field1}, \ref{eq:field2}, \ref{eq:field3} we can produce the optical signals that a model may use as input during real-time operations. \textcolor{black}{The simulation provides a data sampling rate of 2048Hz. This was chosen as it provides a good compromise between high resolution of the trajectories and the optical signals while remaining manageable in data volume for our models to handle long sequence inputs} \GVnew{and in computation speed for the real time implementation of the results.} An example of the motions simulated are in figure \ref{fig:motion_signals}. Together they produce a set of 10 optical signals shown in figure \ref{fig:motion_signals}. 

\textcolor{black}{Notice that in figure \ref{fig:motion_signals} the typical motion range between \(- 2\times 10^{-6} \mathrm{m}\) to \(2\times 10^{-6} \mathrm{m}\) which spans multiple half-wavelengths of \(\frac{\lambda}{2} = 5.32\times 10^{-7} \mathrm{m}\).}

\subsection{Hardware and Engineering Restrictions}
\label{sec:hardware_restriction} %% OK
We also have engineering restrictions at production time. For any technique to be of practical use we need to produce position estimates at approximately 2048 times per second on a single CPU core. This restrictive limitation is due to the fact that LIGO’s realtime system executes control tasks on single processors, which makes parallelization unfeasible \cite{ligo_sys}. Furthermore, there are no GPUs available on the realtime machine.

Beyond just simulating free motion, we \GV{can also simulate the interaction of} the \GV{feedback control loops} in order to demonstrate the locking technique.

\subsection{Problem Statement}
\label{sec:problem }
With these assumptions in place, we can cast the difficult non-linear control problem into an easier state estimator problem. Given the historical data on the optical signals, can we use that to estimate the current state of the mirror's positions? If so can we do this under the constraints of our real-time hardware? We know that if we can answer this question then the control problem is solved, since controlling the motions of pendulums given the state is a well-known and solved problem.

\section{Methods}
\subsection{Our Approach}
\label{sec:approach}
 \textcolor{black}{Here we begin by introducing our approach which will be expanded on in subsequent sections.}
Firstly, as discussed in section \ref{sec:problem } the control problem can be solved if we can reconstruct the mirror's states. However this is difficult because the relationship of input data to states is non-unique and non-linear. \textcolor{black}{ 
 Recall that the non-linear aspect comes from the equations that govern the laser fields (see section \ref{sec:non_linear}), and the non-uniqueness comes from the exponential oscillatory terms}. To address the non-linear aspect,  we have accurate simulations \GVnew{that allow} us to naturally cast the problem into a supervised machine learning problem. \textcolor{black}{ This is because these simulations effectively provide us targets which are the true trajectories, thus the supervised learning objective is to learn a mapping between the optical signals} \GVnew{ and the target trajectories}. To address the non-unique aspect we ''wrap'' the data by taking the trajectories modulo \textcolor{black}{\(\frac{\lambda}{2} \) in the \(Z_1/Z_2\), obtained from MICH and PRCL via a linear transformation that will be explained in section \ref{sec:wrap_data} below}. Using this wrapped data we train a \textcolor{black}{ Gated Recurrent Unit (GRU)} to learn the non-linear mappings between the optical signals and the states. 
 \textcolor{black}{We chose GRU networks firstly for their recurrent architecture \GVnew{and their ability to deal efficiently with} time series. Secondly, we chose the GRU for its superior performance over classical Recurrent Neural Networks on long time series data}. Finally, we trained a model on the wrapped positions, \textcolor{black}{however this introduces artificial discontinuities in the data which need to be fixed to reflect the true motion of the mirrors}. Thus we need to unwrap the data using techniques like the Kalman Filter to provide the best state estimate to ensure continuity in trajectories. \textcolor{black}{Here the Kalman filter is a method to help combine information from different sources to provide the best optimal estimation, using sensor data and knowledge of the systems dynamics. For our use case we would need position and velocity estimates} \GVnew{from the trained GRU network as sensors. It will be sufficient to use } \textcolor{black}{ a simplified model of the mirror's dynamics (constant velocity)}. To utilize the Kalman Filter we \textcolor{black}{ need predictions on both the \textit{position} and the \textit{velocity} along with their \textit{uncertainties} \cite{intro_kalman}}. Hence we will use techniques in probabilistic deep learning to aid our model  \textcolor{black}{in producing uncertainties as well.} \textcolor{black}{Here, Probabilistic Deep Learning is a sub class of deep learning methods built to produce outputs that model probability distributions given input data \cite{probalistic_1}. Examples include Bayesian Neural Networks\cite{bayes_net}, Variational Autoencoders\cite{autoencoder}\cite{vae} and more. }

\subsection{Wrapping Data - Non-Uniqueness}
\label{sec:wrap_data}
As discussed in section \ref{sec:non_linear} a core problem is that there exists an infinite number of \GVnew{positions that would produce the same} optical signals we inputted to the model. \textcolor{black}{Any of these solutions would produce the same optical signal}. \textcolor{black}{ Recall, this is due to the fact that the laser field equations are periodic}. This is not a problem, \textcolor{black}{ choosing any solution would suffice.} The issue is, once a solution is chosen, we need to follow that solution over time to preserve the continuity of \textcolor{black}{the mirror's} trajectories. \GVnew{ To avoid the non-uniqueness problem}, \textcolor{black}{we can ”wrap” the data, but to figure that out we need to find the periodicity of these optical signals and how the oscillatory terms in the laser field equations depend on PRCL and MICH.}

\GV{We inspect the field equations \ref{eq:field1}, \ref{eq:field2}, \ref{eq:field3}: the periodicity is due to the complex exponential terms, that are functions of the variable $\phi_{MICH}$ and $\phi_{PRCL}$ from eq. \ref{eq:phimich}, \ref{eq:phiprcl}. With some straightforward algebraic computations one can show that the only terms appearing in complex exponential are the following lengths degrees of freedom, obtained with an invertible linear transformation from MICH and PRCL:
\begin{equation}
    Z_1 = 2\left(L_{PRCL} + \frac{L_{MICH}}{2}\right)
    \label{eq:z1}
\end{equation}
\begin{equation}
   Z_2 = 2\left(L_{PRCL} - \frac{L_{MICH}}{2}\right)
   \label{eq:z2}
\end{equation}
All fields, and therefore all optical signals, are independently periodic in the two variables $Z_1$ and $Z_2$ with period $\lambda/2$.
}

To wrap the data, firstly we will take the PRCL and MICH positions and follow the transformations described for \(Z_1\) and \(Z_2\). After that is done, we iterate through the data and linearly shift up or shift down by integer multiples of \(\frac{\lambda}{2}\) such that all positions are restricted between \(\frac{\lambda}{4} \to -\frac{\lambda}{4} \). Then we transform back returning the PRCL and MICH positions successfully wrapped. This technique guarantees that for every signal the corresponding wrapped position is unique. \textcolor{black}{The uniqueness allows us to invert the state estimation problem.} 
\\
Once again, we will need to undo this process with the model outputs, since this is was done to create a trainable model and thus are not representative of true motions of the mirrors \textcolor{black}{ at the boundary where we introduced instantaneous jumps from wrapping the data}.

\subsection{Preparing Data}
\label{sec:preprocessing}
\textcolor{black}{Recall that we are working with is a set of 10 optical signals retrieved from our simulation described in \ref{sec:simulate} and \GV{shown as an example} in figure \ref{fig:motion_signals}.} These 10 optical signals need to be normalized, and thus to do so we simulated about 40,000 seconds \textcolor{black}{data} and found the maximum \textcolor{black}{ of the absolute value of each signals} and used it as a \textcolor{black}{normalization constant} throughout all training, testing \textcolor{black}{ and in production}.\textcolor{black}{ The data used to provide the normalization constants is also the training data.}

Next  we prepare data for training and testing the\textsc{position} \textcolor{black}{model}. We first simulate data as described in \ref{sec:simulate} and then wrap the positions using the technique described in section \ref{sec:wrap_data}. We will use these wrapped positions as the targets \textcolor{black}{ to train} our \textsc{position} model. We normalize the positions by \(\frac{\lambda}{4}\) \textcolor{black}{ to bring the target values between \(-1\) and \(1\)}. We simulated about \(40000 \)s of trajectories. We took \(0.5\)s intervals of historical data and asked the model to predict the \textcolor{black}{normalized} wrapped position of the current state at the end of each interval. In total we had \(63839\) training samples and \(15960\) testing samples. \textcolor{black}{ This was the maximum number of samples we could fit both into memory and to train in a reasonable amount of time.}

Then we prepare data for the \textsc{velocity} model. We know that although the position is not unique, the velocity of the mirrors are in fact unique. Thus no need to wrap the data. We obtained the velocity of the mirrors by computing the finite differences of the positions by looking at a \textcolor{black}{current time step \(t\) and one time step a head \(t+1\) or \(\frac{1}{2048}\) seconds ahead.} Also note that the velocity is unbounded and thus we set \textcolor{black}{a normalization constant of} \(2\lambda\) m/s. 

\GVnew{To increse diversity, we also added data produced with the time-domain simulation that has been used later for the lock acquisition tests, as explained in section \ref{sec:simulate}, accounting for about one sixth of the number of examples (95639 training and 23960 test)}

\subsection{Probabilistic Models}
\label{sec:probablistic_model_def}
As mentioned in section \ref{sec:approach} we need a means to estimating uncertainties in our model \GV{predictions}. To do so, we take inspiration from Variational Autoencoders \cite{vae} in our design. Consider a classical feed-forward neural network for a regression problem. Now consider that the last layer splits into two layers, \textcolor{black}{see figure \ref{fig:model_design}}. \textcolor{black}{Recall that we are trying to model a probability distribution. We chose to model it with a Gaussian. } Thus we denote one layer as the mean \(\mu\) and the other as the standard deviation \(\sigma\), \textcolor{black}{ where the \(\mu,\sigma\) are the parameters of a Gaussian distribution}. Notice that this is effectively the encoder model of a Variational Autoencoder \textcolor{black}{with the sampling layer removed}. Then when optimizing the model, we compute the probability using   
\begin{equation}
    P(y | \mu, \sigma^2) = \frac{1}{\sqrt{2\pi \sigma^2}} e^{\frac{1}{2\sigma^2}(-(y-\mu)^2)}
    \label{eq:loss_1}
\end{equation}
which checks how well the target agrees with our predicted probability distribution.  We try to maximize this value.

Intuitively this \textcolor{black}{ could be an effective way} \textcolor{black}{to obtain} the error of the estimator. \textcolor{black}{By conceptually investigating the limiting behaviors we see that if} the model were poor at making predictions, the variance would be high. This means that since the model does not have an accurate estimation of the mean, it compensates that fact by estimating a larger variance this way there is still a random chance that the model predicts a solution close to the target. However, if the model is highly accurate, then the variance should drop and the mean should align with the target value and there is a higher probability the predicted data points fall close to the target. \textcolor{black}{Furthermore, we see that the model would also naturally produce larger uncertainties around discontinuous points. This is because neural networks produce smooth outputs and thus have trouble modeling discontinuous behaviours and therefore to compensate for poor inference the model produces high uncertainty. This is exactly what we wanted since high uncertainty suggests to our Kalman filter to rely more on known dynamics to predict the position rather than the \textsc{position} model.} Overall, in theory as the model improves, the variance should drop and the mean should approach the target value. We can mathematically formulate this as an optimization problem \textcolor{black}{that} minimizes the negative log - likelihood using weights \(\theta\) given \textcolor{black}{neural network} model function \(p\)  and given input data \(x\):

\begin{equation}
p(x;\theta) = \mu, \sigma,\quad 
\min_{\theta} [- \ln(P[y | p(x;\theta)])]
\label{eq:min_log}
\end{equation}

\textcolor{black}{In the end, we frame the optimization problem so that the neural network produces outputs modeled by a Gaussian distribution. Specifically, it forces the model to produce the best estimation of the targets } \GVnew{ and at the same time produce realistic estimates of how close the output is to the target.}

\subsection{Position and Velocity Model}
\textcolor{black}{With this technique of} modeling uncertainties described in \ref{sec:probablistic_model_def} we can develop our \textsc{position} and \textsc{velocity} models. Firstly we know that from testing our existing hardware on a single core of an \textsc{AMD EPYC 7763 64-Core Processor} that a pure C implementation of our neural network has an upper limit of \(\approx 70,000\) parameters \textcolor{black}{ for each model} (both models run on a the same CPU core). \textcolor{black}{We know that this is around the upper limit since, we found that models with 80,000 parameters were too slow while models with 60,000 parameters (all with the same architecture) were faster than required.}

For the \textsc{position} model, \textcolor{black}{it contains 2 GRU layers \cite{chollet2015keras} followed by a Dropout Layer \cite{dropout} then 3 fully connected dense layers with Leaky ReLU activations \cite{leakyrelu}. Lastly it splits to two separate layers with linear activation for the mean and softplus activation \cite{softplus} for the standard deviation model. See table \ref{tab:positionmodel} for specific hyperparameters. This gave a total of 67,827 trainable parameters.}

\begin{table}[tb]
    \centering
    \begin{tabular}{cc}
\hline
Layer & Parameters
\\
\hline
Layer 1 & GRU: 15 units \\
Layer 2 & GRU: 128 units, Dropout: 0.3 \\
Layer 3 & Dense: 64 units, Dropout: 0.3 LeakyRelu: 0.3\\
Layer 4 & Dense: 32 units, Dropout: 0.3 LeakyRelu: 0.3\\
Layer 5 & Dense: 16 units, Dropout: 0.3 LeakyRelu: 0.3\\
Mean & Dense: 2 units,  Linear activation\\
Standard Dev & Dense: 2 units,  Softplus activation\\
\hline\\
\end{tabular}
\caption{Architecture of the \textsc{position} model}
\label{tab:positionmodel}
\end{table}

\textcolor{black}{For the \textsc{velocity} model, the design is identical to the \textsc{position} model except with different hyperparameters. For specifics see table \ref{tab:velocity}. This gave a total of 77,895 trainable parameters. }

\begin{table}[tb]
    \centering
    \begin{tabular}{cc}
\hline
Layer & Parameters
\\
\hline
Layer 1 & GRU: 20 units \\
Layer 2 & GRU: 128 units, Dropout: 0.3 \\
Layer 3 & Dense: 110 units, Dropout: 0.3 LeakyRelu: 0.3\\
Layer 4 & Dense: 32 units, Dropout: 0.3 LeakyRelu: 0.3\\
Layer 5 & Dense: 17 units, Dropout: 0.3 LeakyRelu: 0.3\\
Mean & Dense: 2 units,  Linear activation\\
Standard Dev & Dense: 2 units,  Softplus activation\\
\hline\\
\end{tabular}
\caption{Architecture of the \textsc{velocity} model}
\label{tab:velocity}
\end{table}

\begin{figure}[H]
\includegraphics[width=\linewidth]{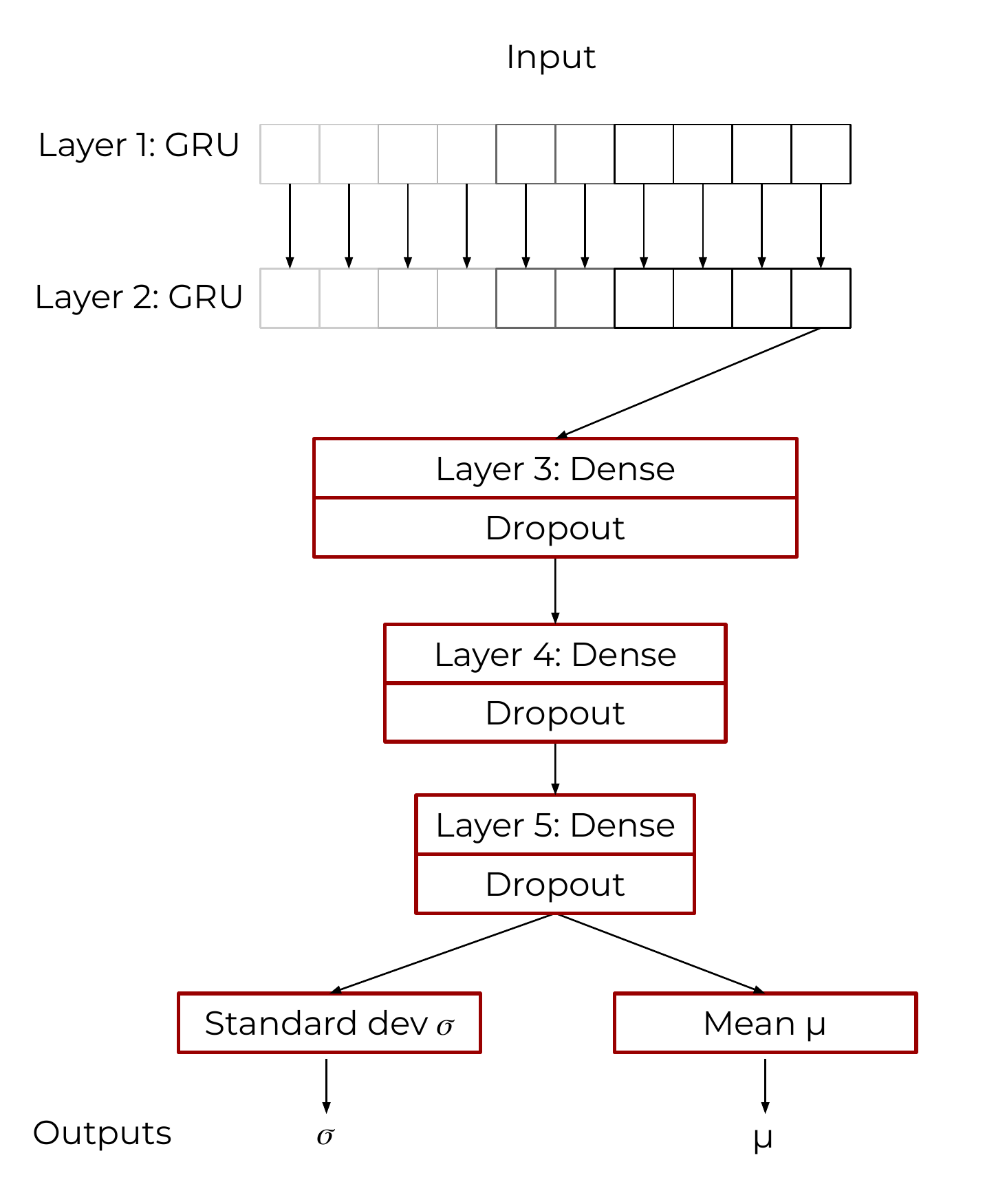}
\caption{\label{fig:model_design} This depict the general structure \GVnew{of the position or velocity models}, for the specific parameters check table \ref{tab:positionmodel} and table \ref{tab:velocity}. } 
\end{figure}

For a visual representation as to how data flows through the model please see figure \ref{fig:model_design}.

\subsection{Training Position and Velocity Model}
We trained both models with 3000 epochs with early stopping with a patience \footnote{Patience: The number of epochs to wait before stopping if there is no progress on the validation set} of 100 epochs using the \textsc{Adam} optimizer \cite{adam}. We noticed improvements in the models training by implementing a simple warm up learning rate plus exponential decay scheduler, see algorithm \ref{alg:scheduler}. We \GV{used} a warm--up period of 20 epochs, with a decay time of 500 epochs, with an initial learning rate of \(1\times 10^{-6}\), a base learning rate of \(1\times 10^{-4}\), and a minimum decayed learning rate of \(1\times 10^{-7}\). \GV{The batch size is 64 for the \textsc{position} model and 500 for the \textsc{velocity} model. Those settings gave the best performing model and are obtained by manually tuning the hyperparameters}. 

\begin{algorithm}[H]
\caption{Learning rate scheduler}\label{alg:scheduler}
\begin{algorithmic}
\Require  \textsc{epoch} is current epoch
\Require  \textsc{lr} is learning rate
\Require  \textsc{warmup\_epochs} is number of epochs to warm up
\Require  \textsc{initial\_lr} is initial learning rate
\Require  \textsc{min\_lr} is minimum learning rate
\Require  \textsc{base\_lr} is learning rate we want to warm up to
\Require  \textsc{decay\_epochs} is number of epochs we wish to take to decay to \GV{the minimum learning rate}.

\If{\textsc{epoch} \(\geq\) \textsc{warmup\_epochs}}
     \State $pct \gets \textsc{epoch} / \textsc{warmup\_epochs}$
    \State RETURN $(( \textsc{base\_lr}-\textsc{initial\_lr}) \cdot pct) +  \textsc{initial\_lr}$
\EndIf

\If{\textsc{epoch} \(<\) \textsc{warmup\_epochs} and \textsc{epoch} \(<\) \textsc{warmup\_epochs}\(+\) \textsc{decay\_epochs}}
     \State $pct \gets 1- \frac{\textsc{epoch}} { \textsc{warmup\_epochs}}$
    \State RETURN $(( \textsc{base\_lr}-\textsc{initial\_lr}) \cdot pct) +  \textsc{min\_lr}$
\EndIf   
\textcolor{black}{
\If{\textsc{\textsc{epoch} \(>\) \textsc{warmup\_epochs}\(+\) \textsc{decay\_epochs}}}
    \State RETURN $\textsc{min\_lr}$
\EndIf    
}
\end{algorithmic}
\end{algorithm}

During our training phase we noticed that, \textcolor{black}{ when strictly optimizing for equation \ref{eq:min_log}}, the reconstructions of the \GV{mean values of the estimates} were relatively poor but with sensible uncertainties . We believe this is in part due to the low capacity of the model. \GV{To enable us to have more} control over how the model weighs the importance in predicting the mean versus estimating the uncertainty we devise a new loss function \(\mathcal{L}_\theta \) shown in equation \ref{eq:loss}.
\begin{equation}
\mathcal{L}_\theta = - \ln[P(y | p(x;\theta)) + \alpha \cdot (y - p(x;\theta)_{\mu})^2]
\label{eq:loss}
\end{equation}
We denote \(p(x;\theta)_{\mu}\) as the \(\mu\) value returned by the model. Simply put, \GV{we} added an additional mean squared error term to help drive the predicted mean closer to the actual position. We \GV{chose} a scaling factor \(\alpha =10\) \GV{such that the model} produce favourable results.

\begin{figure}[H]
\centering
\includegraphics[width=1\linewidth]{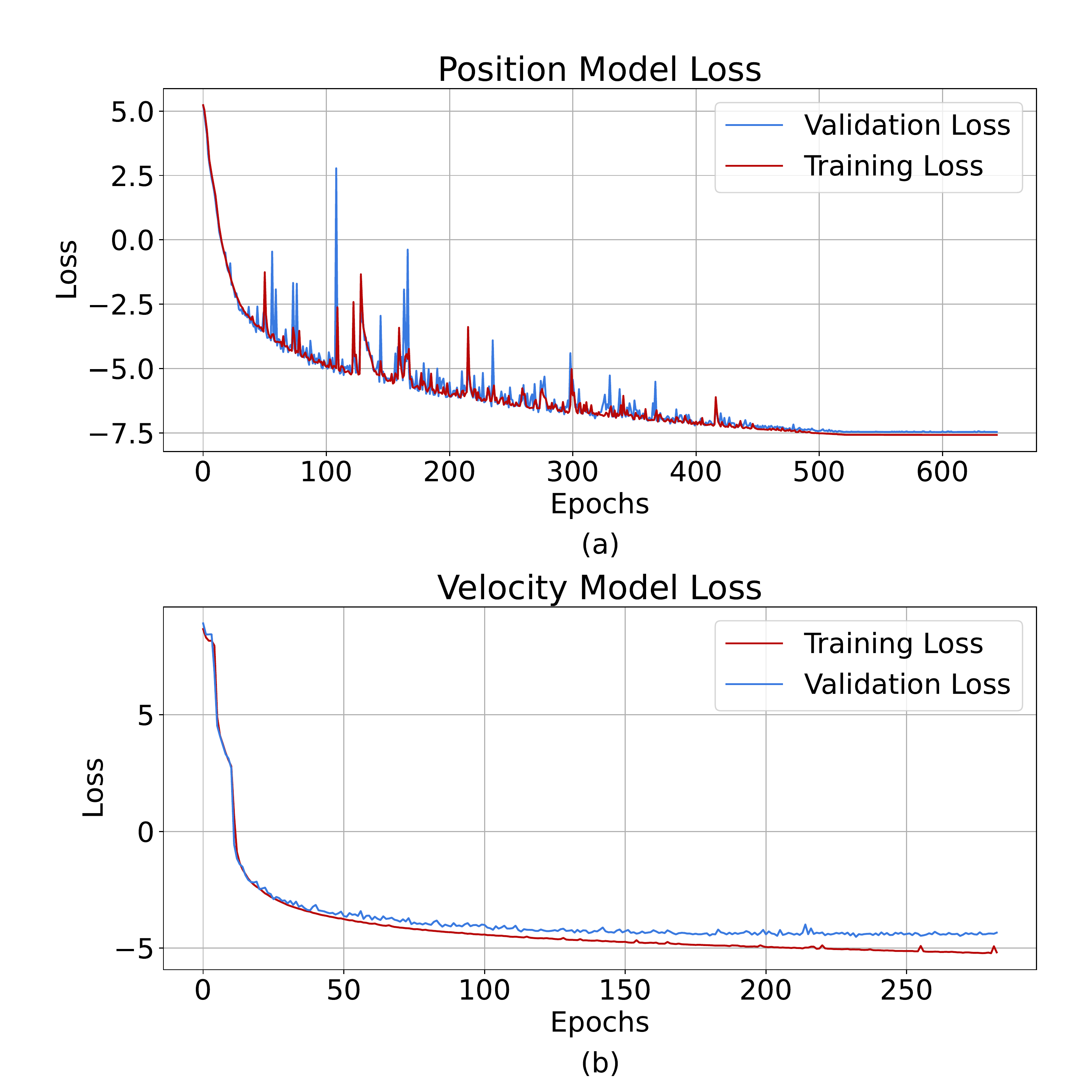}
\caption{\label{fig:loss_curve_pos} (a) Is the loss curve for the position model whereas (b) is the loss of the velocity model. Notice \GVnew{in the} \textsc{velocity} model \GVnew{training}, the relatively smoother loss curve is due to the larger batch size used. } 
\end{figure}

Now we check the training results of both models, specifically their loss curves in figure \ref{fig:loss_curve_pos}.  We see that the position model has plateaued in performance but has yet to completely overfit since the validation and the testing curve remain close  together. On the other hand the velocity model has a harder time training as the validation curve tapers at a much slower rate than the training curve, hence the early stopping preventing the model from training any longer. This indicates to us that stopping training around the times chosen was close to optimal.

\subsection{Testing Model}

\label{sec:testing}

\begin{figure*}[t]
\centering
\includegraphics[width=1\textwidth]{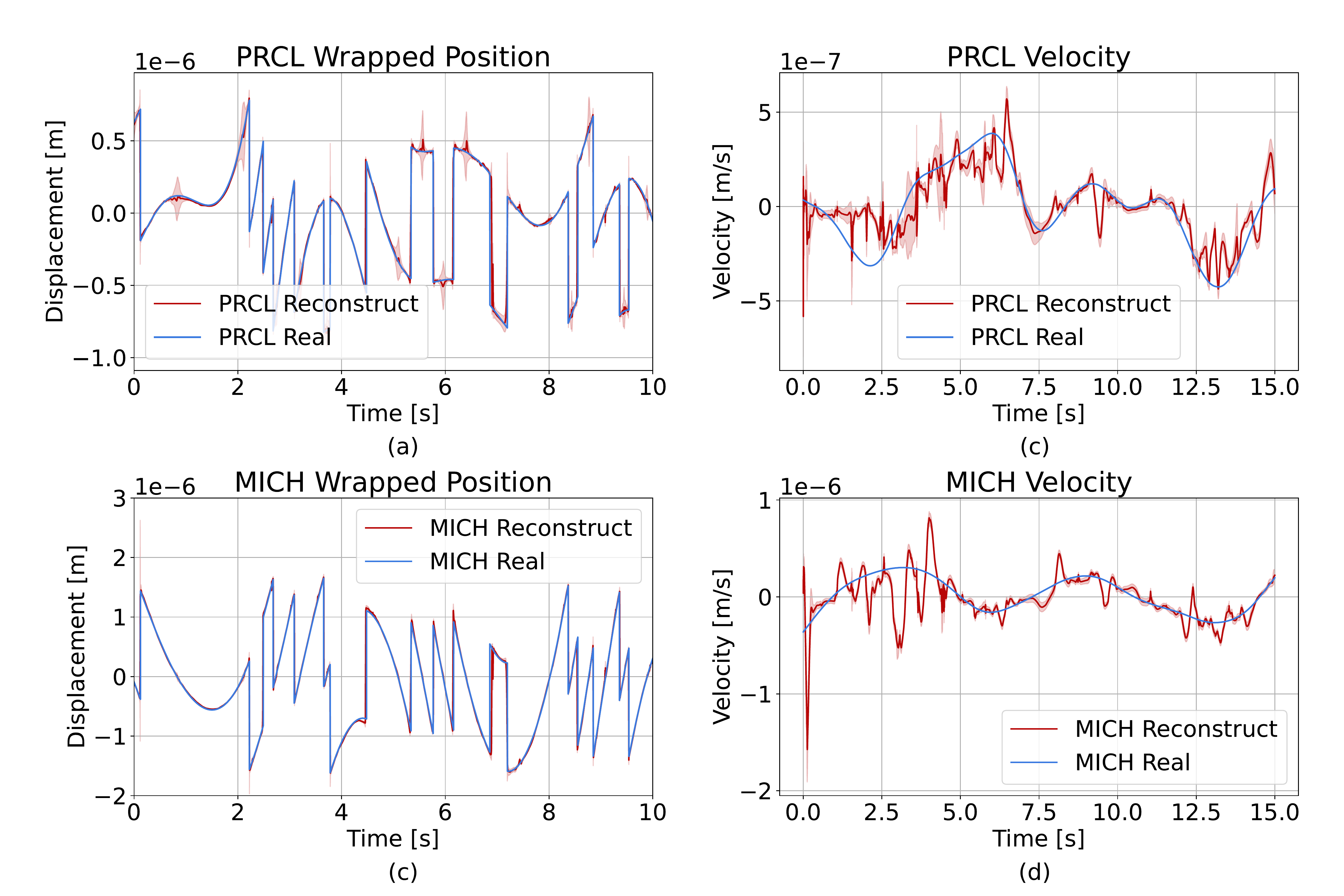}
\caption{\label{fig:reconstruct} Panel (a) \GVnew{shows} the wrapped positions reconstructed by the model of PRCL \GVnew{where the} shaded area \GVnew{shows the uncertainty as predicted by the model}. Similarly for panel (b), \GVnew{showing the PRCL velocity prediction}. In panel (c) - (d) we \GVnew{show the results for MICH}. We notice that the velocity reconstructions are poorer than the position reconstructions.} 
\end{figure*}

We now test the performance of the models by \textcolor{black}{running a new simulation of the mirror motions and optical signals, and running the model with continuous inputs of signals to produce a  10 seconds worth of predicted \textit{wrapped} trajectory. } The \textsc{position} \textcolor{black}{model appears to produce reliable predictions in that predictions match the targets closely whereas the \textsc{velocity} model produces less accurate results. We will show that this performance is sufficient in acquiring the lock in subsequent sections.}   \GVnew{The results are shown in figure} \ref{fig:reconstruct}.

\GV{We note that the velocity model performance is worse than the position model}. \GV{We believe the} reason for \GV{this reduced performance} is the limited capacity of the model, \GVnew{that} thus underfitted. \textcolor{black}{We previously attempted to train with larger models and varying architectures and saw marginal improvements. Thus we kept the smaller model due to the engineering constraints \GVnew{described} in section \ref{sec:general_challenges}. We have also tried training larger models with L1 regularization to remove small weights, however we discovered that the model still contained too many non-vanishing parameters after training.} We attribute these training challenges to the fact that \GV{the} velocity \GV{values are} unbounded whereas the position has bounded values.  \textcolor{black}{Therefore when training a model it is harder to fit the velocity data since the target space is larger and requires more generalization. }This result will affect the design of our Kalman filter, \GVnew{as described below} in section \ref{sec:kalman}.

\subsection{Sensor Fusion - Kalman Filter Inspired}
\label{sec:kalman}
The last part of the solution is to implement a version of the Kalman filter for our special case. \textcolor{black}{ Specifically, we want to use an approach inspired by the Kalman filter to fuse two simple measurements, the position and velocity, to unwrap the data to preserve continuity in predicted trajectories. Our Kalman filter is unique in that our measurements come from neural networks. Since the rate at which new predictions are available is high on the time scale of the state evolution, we can use a simplified dynamical model.}. We begin with building intuition. For every given new time step we can \GVnew{use the models to} predict the wrapped position and \textcolor{black}{the velocities}. There are two things we can do with this data. First we can take the positions and the velocities \GVnew{at the previous time step} and use simple kinematics to estimate the next positions \GVnew{ assuming that the interferometer degrees of freedom evolve as free bodies moving at constant velocity}. It is important to note that the real dynamics of the mirror's positions is not \GVnew{so simple}. This is because the mirrors are subject to random external disturbances due to the ground \GVnew{seismic} motion, and are also suspended to pendulums \GVnew{with resonances at about 1 Hz that introduces long-term slow dynamics{} . But on the time scale of a few time steps the mirror dynamics can be approximated as a constant velocity. }Secondly, we can also get the \GVnew{model prediction for the} wrapped position \GVnew{at the current time step}. \textcolor{black}{However, recall from section \ref{sec:wrap_data} that there are an infinite number of possible solutions which can produce the observed optical signals: our goal is to select the optimal solution with no discontinuities.} Intuitively, since we can use the dynamics to estimate a forward position, we can use this information to help us "select" one of the infinite solutions for the positions: \textcolor{black}{we need to optimally combine the data of the estimated position derived from the dynamics and the estimated position from our sensor (the \textsc{position} model) to produce the best estimate. This takes inspiration from the Kalman filter.}

\begin{figure}
\centering
\includegraphics[width=1\linewidth]{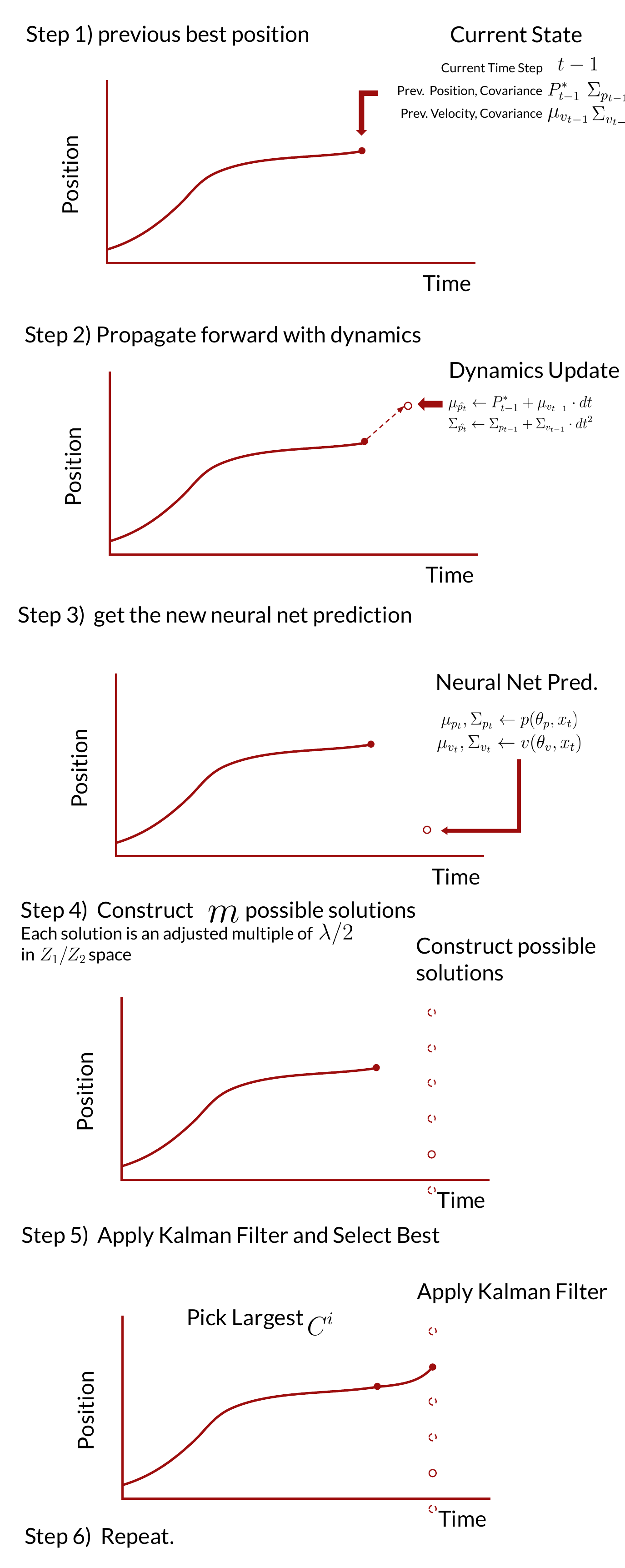}
\caption{\label{fig:algorithm} Here we visualize the algorithm \ref{alg:cap}. 1) start from previous step best position 2) propagate forward with dynamics 3) get the new prediction 4) add suitable multiples of $\lambda/2$ to the new model prediction 5) pick the best.} 
\end{figure}

The process is schematically depicted in figure \ref{fig:algorithm}. First, we take the  estimated position and velocity at the previous time step and use the simplified dynamical model to predict the positions at the current time step. Then we use the \textsc{position} model to predict the  wrapped positions at the current time step and we lay out a set of possible solutions by linearly shifting the wrapped positions by multiples of \(\lambda /2 \) in the $Z_1/Z_2$ space. For each possible solution we will compute the Kalman gain $K$. This is a factor computed from the uncertainties of the dynamic prediction and the uncertainty of the sensor data. More concretely, the Kalman gain gives the coefficient of a linear combination of the two Gaussian distribution of positions (one from the dynamics, one the \textsc{position} model) that minimize the variance of the combined prediction. This factor helps us optimally weigh the adjusted wrapped positions and the positions estimated from dynamics to form new Gaussian distribution for each candidate solution. Now we want to pick the candidate position that gives us the maximum probability. This corresponds to finding the candidate distribution that has the highest peak of the Gaussian in the new fused distribution, computed using the Kalman gain. The detailed algorithm is described in inset \ref{alg:cap}.

Finally let us formalize this in an algorithm which takes inspiration from \cite{kalman}\cite{intro_kalman}, please see pseudocode \ref{alg:cap}, where we defined our algorithm. Let us define the model \(p(\theta_p, x_t)\) for the position and \(v(\theta_v, x_t)\) for velocity where the \(\theta\) are the respective weights. This gives us then the mean and covariance \(\mu_{p_t}, \Sigma_{p_t}\) and \(\mu_{v_t}, \Sigma_{v_t}\) where the subscript denotes either position or velocity at times \(t\). \(x_t\) is the optical signals received at time step \(t\).  The total time is \(T\). We denote \(P^*\) as the best prediction of position. \(\hat{p_t}\) is the position predicted by the dynamics. \(dt\) is the time step. 
\begin{algorithm}
\caption{Kalman Filter Inspired Model}\label{alg:cap}
\begin{algorithmic}
% \Require $n \geq 0$
% \Ensure $y = x^n$
\State Initialize: we set $m$ as the number of solutions we search for. \textcolor{black}{This number specifies the how many different adjustments made by shifting by \(\lambda/2\) in the \(Z_1/Z_2\) space do we check for in our algorithm.}
\State Initial estimate
\State $\mu_{p_0}, \Sigma_{p_0} \gets p(\theta_p, x_0)$
\State $\mu_{v_0}, \Sigma_{v_0} \gets v(\theta_v, x_0)$
\State Best estimate initially comes directly from the model.
\State $P^*_0 \gets  \mu_p$
\While{$t \in \{1,2\cdots T\}$}
    \State We first use the dynamics and update 
    \State $\mu_{\hat{p_{t}}} \gets P^*_{t-1} + \mu_{v_{t-1}}  \cdot dt$
    \State $\Sigma_{\hat{p_{t}}} \gets \Sigma_{p_{t-1}} + \Sigma_{v_{t-1}}  \cdot dt^2$
    \State Now we construct a Gaussian mixture of solutions.
    \State Get wrapped position from the model that looks one timestep ahead 
    \State $\mu_{p_{t}}, \Sigma_{p_{t}} \gets p(\theta_p, x_{t})$
    \State $\mu_{v_{t}}, \Sigma_{v_{t}} \gets v(\theta_v, x_{t})$
    \State{ We construct \(m\) possible solutions which have the same covariance but have means \textcolor{black}{shifted by integer multiples of \(\frac{\lambda}{2}\) in the \(Z_1/Z_2\)  space.}  We denote \(\mu_{p_{t}}^i, \forall i\in (0,1,\cdots m)\) \textcolor{black}{ as these adjusted means where $\mu_{\hat{p_{t}}}^i$ is computed by transforming it into the \(Z_1/Z_2\)  space and applying integer multiple adjustments by \(\lambda/2\) and then transforming it back to PRCL MICH space.}}
    \State Compute Kalman Filter Propagation 
    \While{$i< m$}
        \State \textcolor{black}{ Compute the Kalman gain explained in \ref{sec:kalman}}
        \State \(K \gets \frac{\Sigma_{\hat{p_{t}}} }{ \Sigma_{\hat{p_{t}}} + \Sigma_{p_{t}}  } \)
        
        \State \( \mu_{p_{t}}^i \gets  \mu_{\hat{p_{t}}} + K (\mu_{p_{t}}^i- \mu_{\hat{p_{t}}})
        \)
        \State \(\Sigma{p_{t}}^i \gets ( \Sigma{\hat{p_{t}}} + K (\Sigma_{p_{t}} - \Sigma_{\hat{p_{t}}})  ) K^T  \)

        % \State \(\Sigma_{t+1}^i \gets ( \Sigma_{p_{t+1}}^{-1} +\Sigma_{\hat{p_{t+1}}}^{-1})^{-1} \)
        \State \textcolor{black}{Compute the peak value of the Gaussian distribution. }
        \State \(A \gets (\sqrt{\text{det}(2\pi ( \Sigma_{p_{t}}
        +\Sigma_{\hat{p_{t}}}))})^{-1}\)
        \State \(C^i \gets 
         A \cdot\text{exp}[\frac{-1}{2} (\mu_{p_{t}}^i - \mu_{\hat{p_{t}}})^T ( \Sigma_{p_{t}}
        +\Sigma_{\hat{p_{t}}})^{-1}(\mu_{p_{t}}^i - \mu_{\hat{p_{t}}}) ]\)
        \State \textcolor{black}{We use these \(C^i\)'s to find the most likely new distribution.}
     \EndWhile

\State Sort for the \(i\) that maximizes \(C^i\). Denote the best value as \(i^*\)
\State \(P_{t}^* \gets \mu_{p_{t}}^{i^*} \)
\State Now we update the positions
\State \(\Sigma_{p_{t}}\gets \Sigma{p_{t}}^{i^*} \)
\State Update the velocity with the predictions from before.

\EndWhile
\end{algorithmic}
\end{algorithm}

When adapting the Kalman filter for these neural networks we found that the \textsc{velocity} model continuously produces relatively poorer reconstructions than the position model. This is anticipated since during testing phase (see section \ref{sec:testing}) we realized that the velocity model needs to map to a much larger range and thus generalization is more difficult than the position model. However due to parameter constraints this model was still the best performing out of all the experiments. Thus to mitigate the generalization error, in the Kalman filter we artificially scaled the uncertainty of the velocity to be $15$x greater than its \textcolor{black}{original estimate}.  Although this biases against the velocity model trusting more the position model we can justify this since we know that uncertainties of the velocity model is less trustworthy due to our underfitted explained previously. 

\begin{figure}[H]
\centering
\includegraphics[width=1\linewidth]{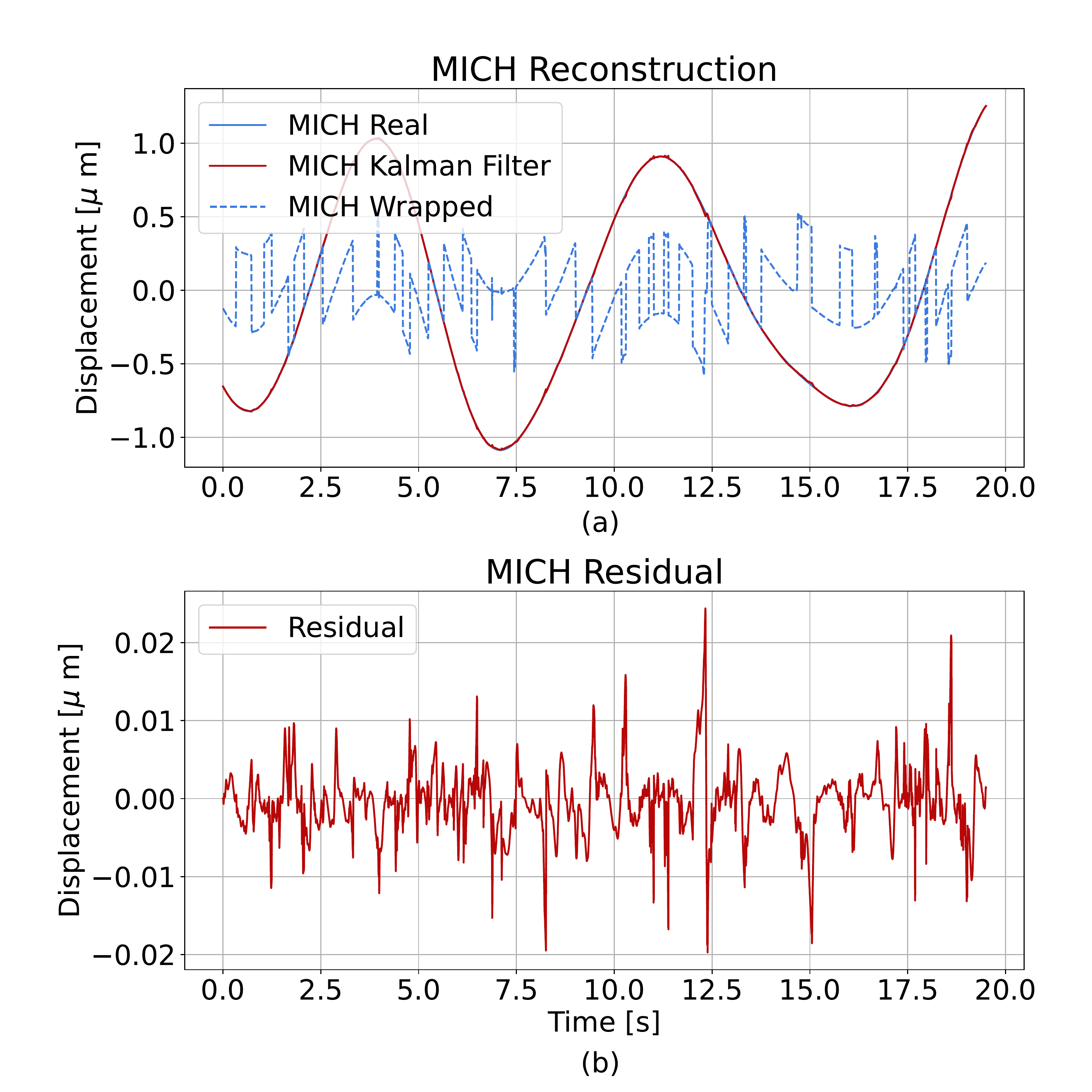}  
\caption{\label{fig:smaller_mich_Kalman} Panel (a) shows \GVnew{the MICH position prediction obtained with the } Kalman unwrapping. Notice that the two trajectories are effectively the same. Panel (b) shows the residuals between the true motions and the reconstructions. We see that the deviations are small, on the order of \(1\times 10^{-8}m\) from zero. } 
\end{figure}

\begin{figure}[H]
\includegraphics[width=1\linewidth]{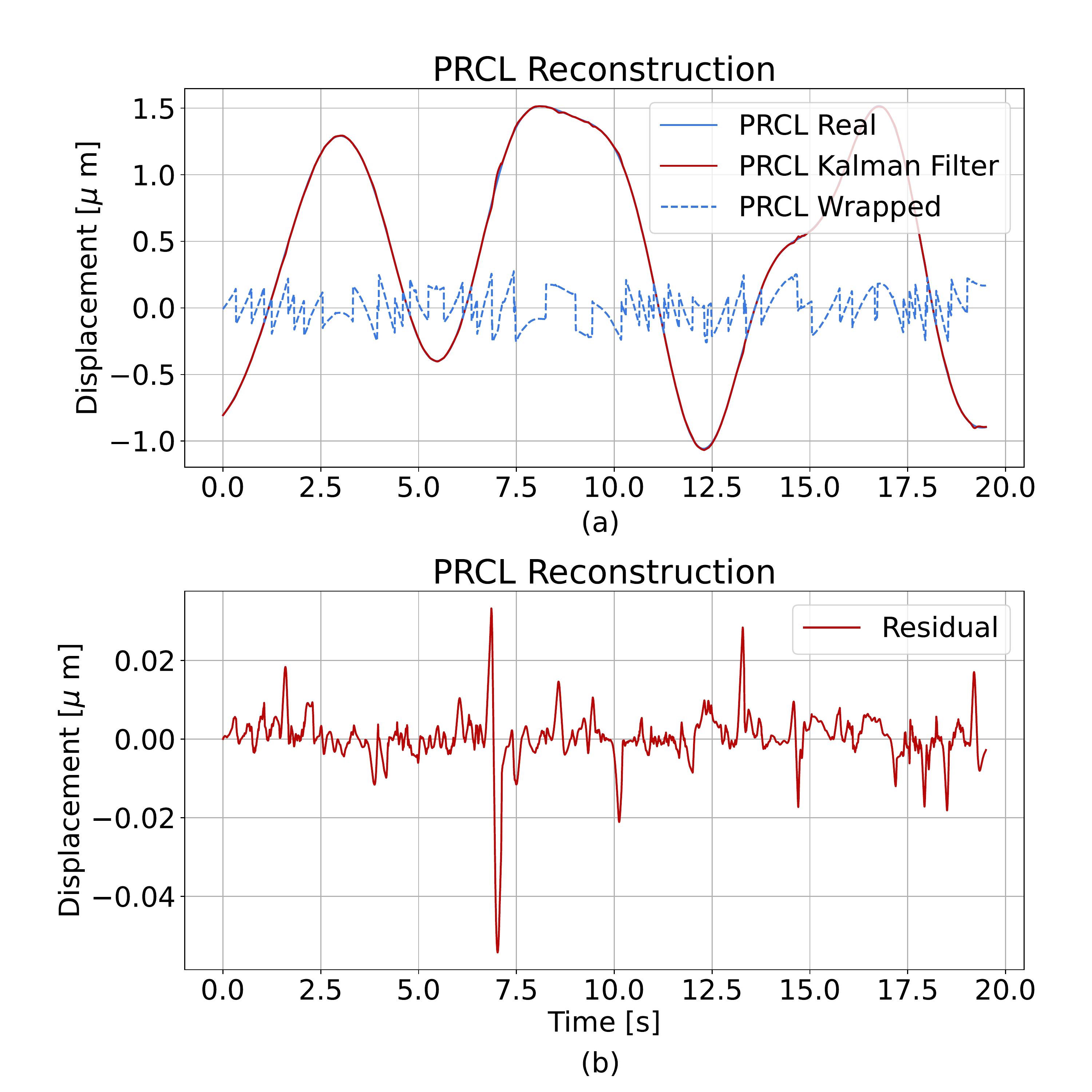}
\caption{\label{fig:smaller_prcl_Kalman} Panel (a) shows \GVnew{the PRCL position prediction obtained with the } Kalman unwrapping. Notice that the two trajectories are effectively the same. Panel (b) shows the residuals between the true motions and the reconstructions. We see that the deviations are small, on the order of \(1\times 10^{-8}m\) from zero. } 
\end{figure}

 \textcolor{black}{We now attempt to reconstruct the mirror's states using this technique we \GVnew{just described}.}   We see that upon first glance our reconstruction appears accurate, \textcolor{black}{see figures \ref{fig:smaller_mich_Kalman} and \ref{fig:smaller_prcl_Kalman}.} To verify such a claim we compute the residuals, \textcolor{black}{ which are the differences between ground truth and estimated values.}. We know that if we are successful then the residuals should be constant \textcolor{black}{ with a value that is one of the integer multiples of \(\lambda/2\) in the $Z_1/Z_2$ space. This is because, \GVnew{as explained} in section \ref{sec:wrap_data}, any of these position solutions would suffice as long as trajectories remain continuous. }\textcolor{black}{We see that our model produces such a result with a constant shift that is allowed by our wrapping technique  with deviations from constancy \GVnew{of the order of} \(10\)nm.} \GVnew{To understand if this accuracy is sufficient, we can compare it with the line-width of the power recycling cavity, that is the maximum displacement from resonance of the PRCL degree of freedom that maintain the power level at more than half the maximum, and ensure a reasonable linearity of the optical signals. If the estimate is accurate within this level, then a control system based on the state estimates would be able to drive the system near resonance, where a classical feedback controller based on linearized optical signals would work. With the parameters of the system simulated here, the power recycling cavity line-width is about 7 nm, of the same order of magnitude of the residual shown in figure \ref{fig:smaller_prcl_Kalman}. We therefore conclude that the performance of out state estimator is sufficient for our goals.}  
 
\section{Results}

\textcolor{black}{Having developed a reliable means of state reconstruction, we can \GVnew{now} address the original question of how to use ML to acquire the lock: \GVnew{that is to design a control system that can bring the system reliably and deterministically to the desired operating point, that is the $(0,0)$ position in the MICH and PRCL space}. In the following sections we demonstrate \GVnew{how} our novel technique successfully solves the locking acquisition problem for the PRMI configuration, in simulation.}

\subsection{\GV{Lock acquisition}}

\begin{figure}
\includegraphics[width=\linewidth]{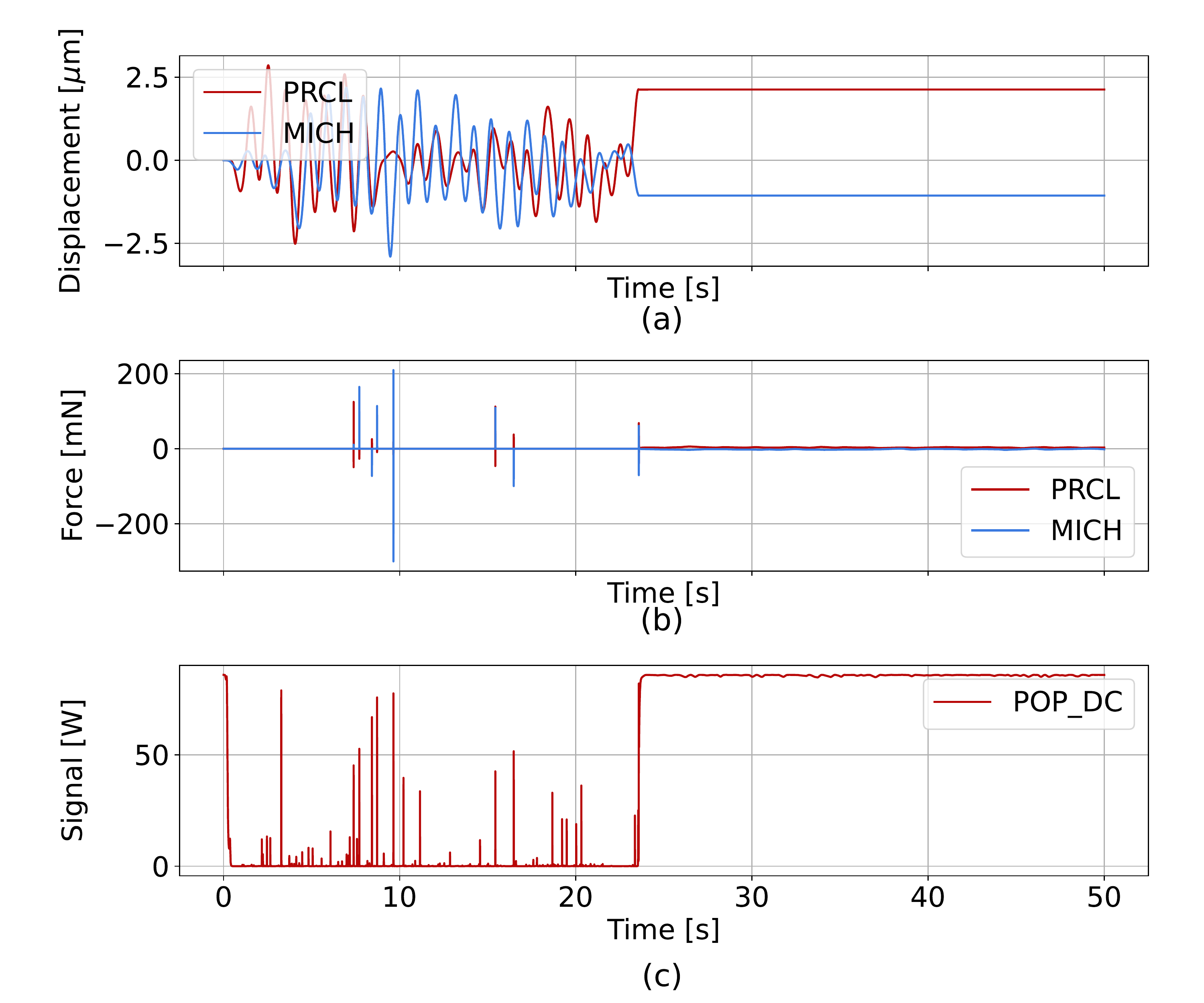}
\caption{\label{fig:classical-lock}Classical lock, here we see three panels depicting a successful classical lock \GV{acquisition} currently used in production. Firstly in panel (a) we see that the motions of the mirrors are driven down such that the optical laser is in resonances, this is shown in panel (c) where the power remains high. In figure (b) we see the forces applied in the PRCL and MICH.} 
\end{figure}

\GV{The main goal of this work was to develop a state estimator for the simulated power-recycled interferometer, and then use it to drive the mirrors to suppress their random seismic motion and maintain the system close to the zero working position. This procedure, called lock acquisition, is currently implemented in Advanced LIGO by relying on a empirical approach: linear feedback controller are engaged when the power measured by the power-recycling-cavity pick-off (POP) exceed a given threshold, typically corresponding to 50\% of the peak value, that is the steady state power measured when the system is maintained at the operating point. An example \GVnew{of the result of this algorithm applied to our simulation} is shown in figure \ref{fig:classical-lock}. } \GVnew{This algorithm is intrinsically non--deterministic, since it relies on waiting for the system to cross near the operating point: when that happens, as detected by large power in the recycling cavity, linear controllers are switched on to quickly stop the mirrors motion and drive them to the working point. This is a process that requires a relatively large force (note that the peak force applied to the mirrors in our simulated lock acquisition is 200 mN, as shown in figure \ref{fig:classical-lock}) and also is not always successful, resulting in multiple attempts and longer times to acquire the lock, of the order of tens of seconds even in the simple PRMI configuration considered here. Finally, the parameters of this algorithm, such as the loop gains and the triggering levels, need to be adjusted manually by experts, to match the system. In more complex configurations, like the dual-recycled interferometer used in Advanced LIGO \cite{ligo_sys}, the increased complexity results in reduced reliability and much longer wait times.}

\GV{The alternative lock acquisition we present here, based on the non-linear state estimator we developed, can overcome some of the limitations of the classical lock acquisition: it can be developed with supervised learning for any optical configuration, without the need of expert knowledge; it does not rely on the system to randomly move close enough to the working point to trigger the controller, and therefore it is faster and deterministic; the full knowledge of the system state allows us to implement feedback control system that can operate with lower peak force, since there is no need to stop the mirror motion in the short time the system spend near resonance, when moving randomly due to seismic ground motion.}

\begin{figure}[H]
\includegraphics[width=\linewidth]{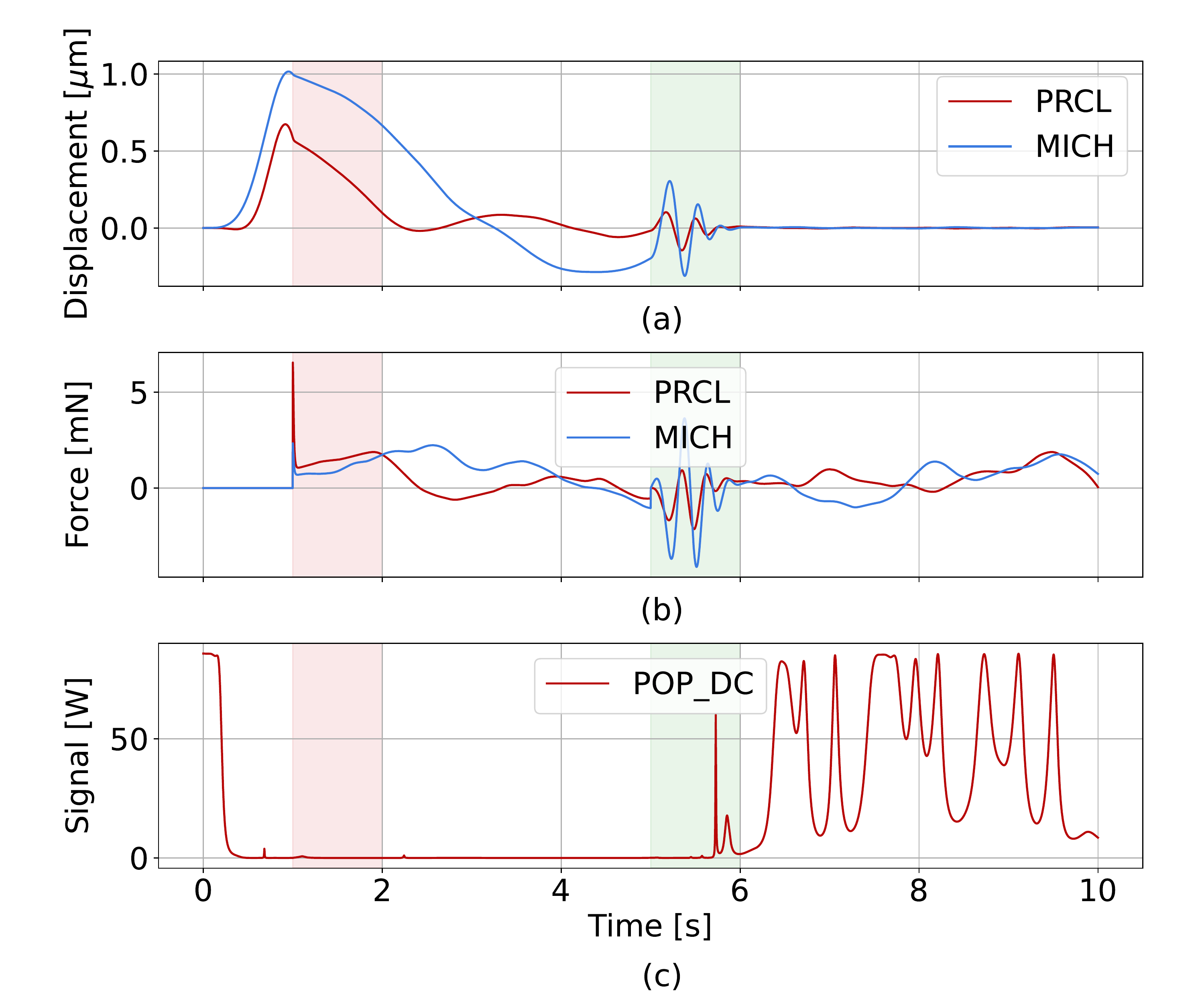}
\caption{\label{fig:locking_scheme} Here we see three panels depicting a successful lock \GV{acquisition} using perfect state estimates. Firstly in panel (a) we see that the motions of the mirrors are driven down to close to 0. Secondly, we show in panel (b) that the force needed to acquire the lock is low. Finally, we see in panel (c) that where the power \GV{in the power recycling cavity} is still fluctuating, but remains high most of the time. This is a good sign since this means the cavities are \GV{maintained near}. Note that the \GV{red shaded region shows the time interval when we ramp up the velocity-damping controller, and the green shaded region shows the time interval when we ramp up the integrator feedback control.}} 
\end{figure}

\GV{We first develop the linear feedback controllers that are needed to drive the system to resonance, assuming perfect knowledge of the system state. We implement a two-step approach: first we engage a velocity-damping feedback loop, that reduces the mirror velocities; then we engage an integrator so that the actual mirror relative positions are driven to zero, that is the cavity resonance.} An example of a completely successful lock \GV{is shown} in figure  \ref{fig:locking_scheme}. \GV{Once the final state shown in figure \ref{fig:locking_scheme} is obtained, the system is actively maintained very near the operating position. In this state it would be possible to engage directly the classical linear feedback controller used for the example shown in figure \ref{fig:classical-lock}, without any triggering, and recover the highly accurate controlled state needed too operate a gravitational-wave detector. This last transition is not shown in our simulations.}

\GVnew{As a second step, we utilize the state estimates described in our methods section to \textcolor{black}{actively} control the motions of the mirrors \textcolor{black}{in numerical simulations}}, \GV{using the feedback control strategy described above}. \GVnew{In this configuration the entire lock acquisition procedure is based on the non-linear state estimator that we developed and the simple and robust feedback controllers just described.}

\GV{Figure \ref{fig:locking_scheme} shows that the estimate produced by our model are good enough to achieve a lock acquisition and maintain the system near resonance.} \GVnew{The accuracy of the control is lower than what is obtained with the classical linear controllers, but as already mentioned, it would be possible to engage them, without triggering, once our control strategy has driven the system near the working point. Our new lock acquisition strategy is completely deterministic, faster than the classical lock acquisition, and requires a much lower peak force on the mirrors (compare the maximum force of 20 mN in figure \ref{fig:locking_scheme}  with the 200 mN in figure \ref{fig:classical-lock}.}

\begin{figure}[H]
\includegraphics[width=\linewidth]{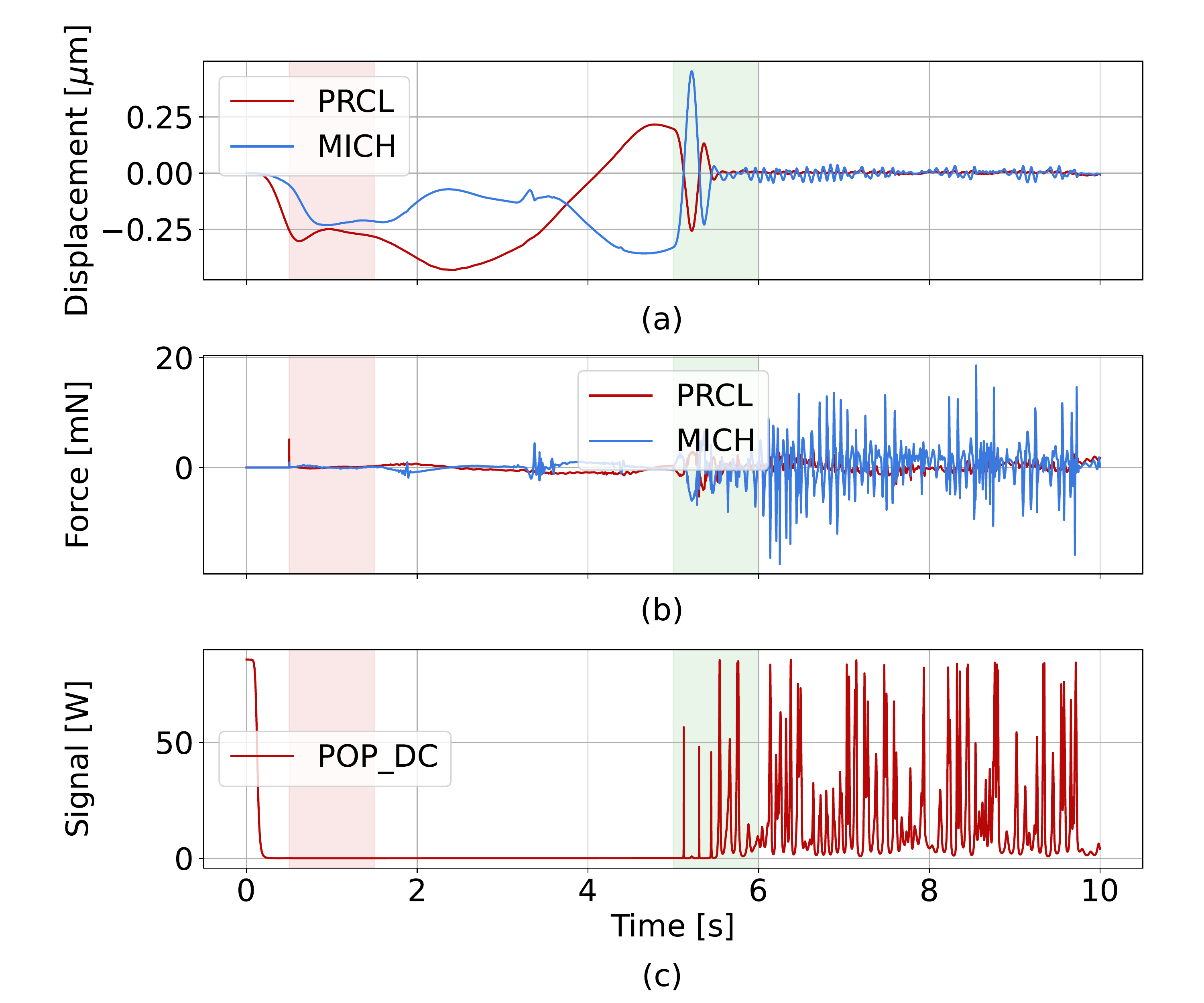}
\caption{\label{fig:locking_neuralnet} Here we see three panels depicting a successful lock \GV{acquisition} using our neural network + Kalman filter state estimates. Firstly in panel (a) we see that the motions of the mirrors are driven down to close to 0. Secondly we see panel (c) where the power in the power recycling cavity is still fluctuating, but remains high most of the time. This is similar to the locking scheme using perfect knowledge of the state thus demonstrating success.} 
\end{figure}

\section{Discussions}

\textcolor{black}{In conclusion, we demonstrated in sections  \ref{sec:approach} that we have successfully developed an algorithm to accurately produce non-linear state estimation of the mirrors positions. Then we proved that this technique is accurate enough to allow acquiring the lock of the Power Recycled Michelson interferometer in a simulation, by using a two stage dampening locking scheme. We believe this is a superior approach to the classical locking scheme currently deployed because it does not rely on expert knowledge about the system. Due to this advancement this opens up possibilities to generalize to many kinds of GW detector configurations for faster development. Furthermore, perhaps most importantly, this technique allows us to lock the motions of the mirrors faster and with more reliability than relying on random motions as in the classical locking scheme. }

\textcolor{black}{Our next step is to implement this on real hardware to verify that this work can translate to the real setting. Should the results not translate, there are a number of investigations that can be done. We can check how well the locking simulation aligns with the real configuration, we can check how well the simulated mirror motions align with the real motions of the mirrors, and lastly we can investigate improvements to the neural network models in sections \ref{sec:testing}, perhaps using larger models or different architectures. For future investigations we could explore if our techniques applied to more complex mirror configurations.}

\section*{Acknowledgment}

The authors would like to thank the kind support of the Caltech SURF program for facilitating this summer project. 

\GV{This material is based upon work supported by NSF's LIGO Laboratory which is a major facility fully funded by the National Science Foundation. The authors gratefully acknowledge the support of the United States National Science Foundation (NSF) for the construction and operation of the LIGO Laboratory and Advanced LIGO as well as the Science and Technology Facilities Council (STFC) of the United Kingdom, and the Max-Planck-Society (MPS) for support of the construction of Advanced LIGO. Additional support for Advanced LIGO was provided by the Australian Research Council. LIGO was constructed by the California Institute of Technology and Massachusetts Institute of Technology with funding from the National Science Foundation, and operates under cooperative agreement PHY-1764464. Advanced LIGO was built under award PHY-0823459. The authors are grateful for computational resources provided by the LIGO Laboratory and supported by the National Science Foundation Grants PHY-0757058 and PHY-0823459. This paper has LIGO document number P2300040.}

\ifCLASSOPTIONcaptionsoff
  \newpage
\fi

\bibliographystyle{naturemag}
\bibliography{references.bib}

\begin{IEEEbiography}[{\includegraphics[width=1in,height=2.25in,clip,keepaspectratio]{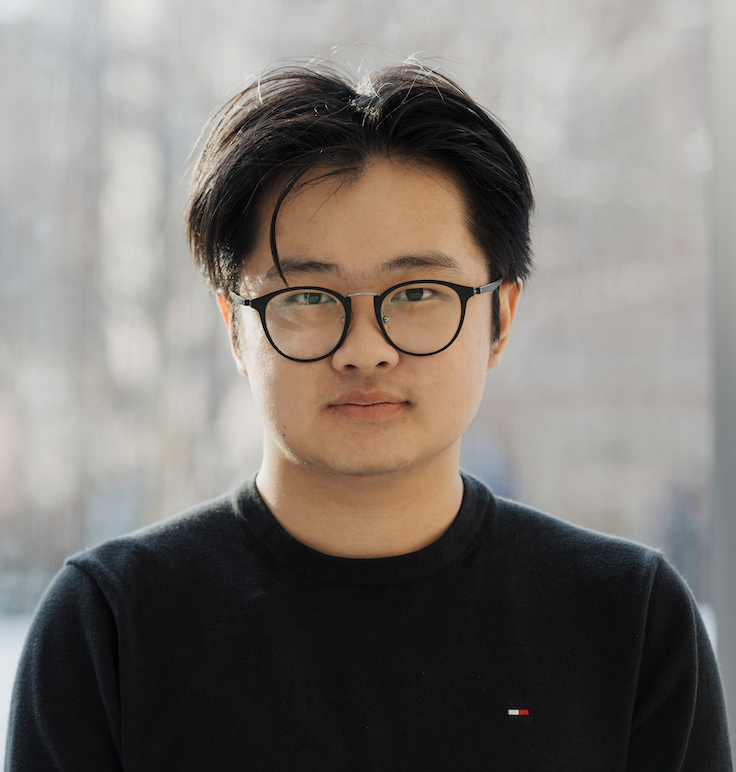}}]{Peter Xiangyuan Ma}
is an undergraduate student at the University of Toronto and Laidlaw Scholar studying Mathematics and Physics. He is curently at UC Berkeley investigating deep learning techniques for radio astronomy. Previously he was an intern at Caltech LIGO working on this current project. Before that he was at the Dunlap Institute of Astronomy and Astrophysics developing machine learning techniques for Fast Radio Burst detection with the CHIME/FRB project. He is interested in leveraging machine learning to accelerate scientific discovery. 
\end{IEEEbiography}

\begin{IEEEbiography}[{\includegraphics[width=1in,height=2.25in,clip,keepaspectratio]{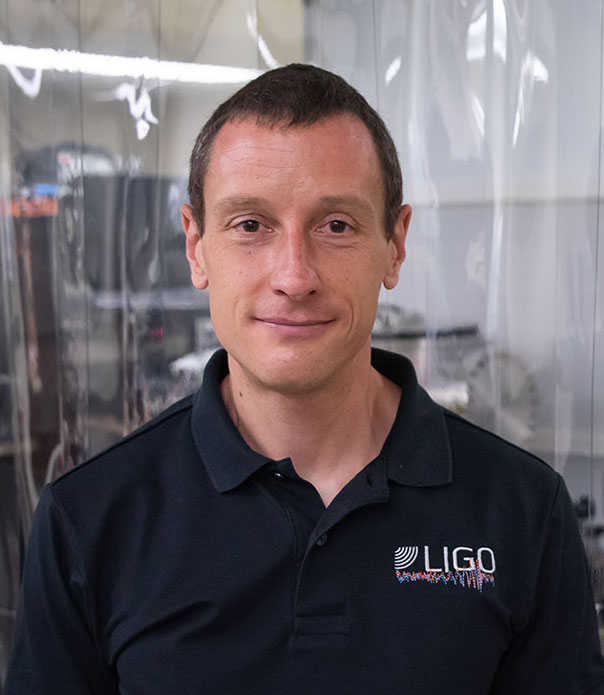}}]{Gabriele Vajente}
\GV{is the Deputy Head of System Science and Engineering of the LIGO Laboratory at the California Institute of Technology. He is currently involved in the development, design, construction and operation of the Advanced LIGO detectors, and on the research and development of future upgrades for gravitational-wave interferometric observatories. Among other research interests, he is studying application of Machine Learning techniques to control and noise reduction in Advanced LIGO and beyond.}
\end{IEEEbiography}

\end{document}